\documentclass[letterpaper, 10 pt, conference]{ieeeconf}
\IEEEoverridecommandlockouts
\usepackage{easyReview} 
\usepackage{algorithmic}
\usepackage{array}
\usepackage{textcomp}
\usepackage{stfloats}
\usepackage{url}
\usepackage{verbatim}
\usepackage{balance}
\usepackage{diagbox}
\usepackage{amsmath,amssymb,amsfonts}
\usepackage{amsmath}
\usepackage{graphicx}
\usepackage{textcomp}
\usepackage{multirow}
\usepackage{makecell}
\usepackage{booktabs}
\usepackage{soul,color}
\usepackage{CJKutf8}
\usepackage{threeparttable}
\usepackage{caption}
\usepackage{subfigure} 

\newlength{\whilewidth}
\usepackage{cite}
\usepackage{algorithm}

\usepackage{graphicx}
\usepackage{etoolbox}

\let\oldtwocolumn\twocolumn
\renewcommand\twocolumn[1][]{%
	\oldtwocolumn[{#1}{
		\begin{center}
			\vspace{-1.5mm}
			\includegraphics[width=\textwidth]{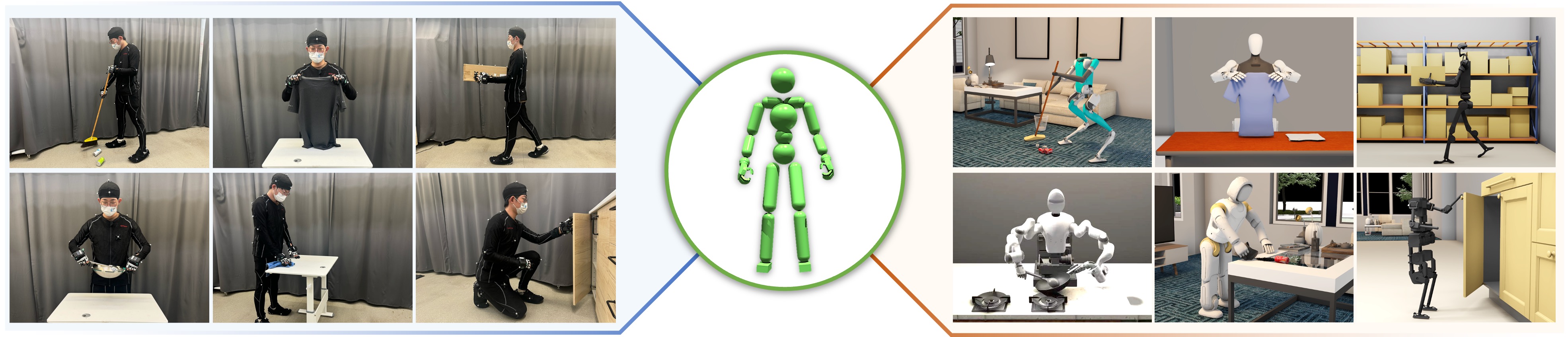}
			\captionsetup{font={small}}
			\captionof{figure}{Human can serve as the prototype of diverse humanoid robots, efficiently learning generalized loco-manipulation skills without considering configuration differences. The framework introduces a unified digital human (UDH) model (the green humanoid) to aggregate human behaviors from embodiment demonstrations. Loco-manipulation skills are generated by combining behavior primitives through adversarial imitation, and subsequently deployed on diverse humanoid robots using efficient motion retargeting and fine-tuning techniques.}
			
		\label{cover}
	\end{center}
}]
}

\begin{document}
\title{\LARGE \bf
	Human-Humanoid Robots Cross-Embodiment  Behavior-Skill Transfer Using Decomposed Adversarial Learning from Demonstration
}
	%
	%
%

\author{Junjia Liu$^{1}$, Zhuo Li$^{1}$, Minghao Yu$^{1}$, Zhipeng Dong$^{1}$,\\ Sylvain Calinon$^{2}$,  \IEEEmembership{Member, IEEE}, Darwin Caldwell$^{3}$,  \IEEEmembership{Fellow, IEEE} and Fei Chen$ ^1$, \IEEEmembership{Senior Member, IEEE}
	\thanks{This work is supported in part by the Research Grants Council of the Government of the Hong Kong SAR via the Grant 24209021, 14222722, 14211723, 14213324, C7100-22GF and in part by the InnoHK of the Government of the Hong Kong SAR via the Hong Kong Centre for Logistics Robotics. (Corresponding
		author: Fei Chen.)}
	\thanks{$^{1}$Junjia Liu, Zhuo Li, Minghao Yu, Zhipeng Dong, and Fei Chen are with the Department of Mechanical and Automation Engineering, T-Stone Robotics Institute, The Chinese University of Hong Kong, Hong Kong SAR (e-mail: {\tt\footnotesize jjliu@mae.cuhk.edu.hk, zli@mae.cuhk.edu.hk, mhyu@mae.cuhk.edu.hk, zhipengdong@cuhk.edu.hk, f.chen@ieee.org}).}
	\thanks{$^{2}$Sylvain Calinon is with the Idiap Research Institute, Martigny, Switzerland  (e-mail: {\tt\footnotesize sylvain.calinon@idiap.ch}).}
	\thanks{$^{3}$Darwin Caldwell is with the Department of Advanced Robotics, Istituto Italiano di Tecnologia, Genoa, Italy (e-mail: {\tt\footnotesize  darwin.caldwell@iit.it}).}
}

\maketitle
\begin{abstract}

	Humanoid robots are envisioned as embodied intelligent agents capable of performing a wide range of human-level loco-manipulation tasks, particularly in scenarios requiring strenuous and repetitive labor. However, learning these skills is challenging due to the high degrees of freedom of humanoid robots, and collecting sufficient training data for humanoid is a laborious process. Given the rapid introduction of new humanoid platforms,  a cross-embodiment framework that allows generalizable skill transfer is becoming increasingly critical. To address this, we propose a transferable framework that reduces the data bottleneck by using a unified digital human model as a common prototype and bypassing the need for re-training on every new robot platform. The model learns behavior primitives from human demonstrations through adversarial imitation, and the complex robot structures are decomposed into functional components, each trained independently and dynamically coordinated. Task generalization is achieved through a human-object interaction graph, and skills are transferred to different robots via embodiment-specific kinematic motion retargeting and dynamic fine-tuning. Our framework is validated on five humanoid robots with diverse configurations, demonstrating stable loco-manipulation and highlighting its effectiveness in reducing data requirements and increasing the efficiency of skill transfer across platforms.
		
		

\end{abstract}

\section{Introduction}

Humanoid robots are increasingly expected to perform human-level loco-manipulation tasks. In recent years, the development of diverse humanoid robot platforms has expanded rapidly, each featuring unique configurations and dynamic properties. While hardware has advanced rapidly, humanoid skill learning remains a significant challenge. Traditionally, loco-manipulation tasks have been divided into sub-objectives like collision avoidance, balance maintenance, and robustness to external forces. While effective, this approach is time-consuming and impractical for the fast-paced development cycles of modern robotics \cite{sleiman2023versatile}. Learning-based approaches for robot skill acquisition are now mainstream due to their flexibility and adaptability. However, the high degrees of freedom (DoFs) in humanoid robots result in a large and complex action space, making it challenging for learning-based algorithms to acquire coordinated skills. Additionally, the scarcity of quality training data hinders progress, as collecting diverse robot data is labor-intensive and time-consuming, especially for loco-manipulation tasks.

Skill transfer has emerged as a critical capability for addressing these limitations. By reusing and adapting learned skills from human demonstrations to robots and across different humanoid platforms, skill transfer reduces the need for extensive retraining and facilitates efficient skill acquisition. However, humanoid skill transfer faces two key challenges. First, \textbf{configuration differences} arise because humanoid robots typically feature joint structures, physical dimensions, and dynamic properties that differ significantly from human embodiments and from one another. These differences make direct skill transfer inefficient and often lead to unstable motion. Second, \textbf{high-dimensional coordination} poses a challenge, as humanoid robots must execute whole-body motions across a high-dimensional action space, requiring coordination of multiple functional parts (e.g., legs, arms, and hands). Addressing these challenges is critical for enabling humanoid robots to learn and generalize complex skills efficiently. Moreover, the increasing diversity of humanoid platforms necessitates a cross-embodiment framework capable of transferring skills between robots without extensive retraining for each new embodiment.

\begin{figure*}[t]
	\centering
	\includegraphics[width=0.9\linewidth]{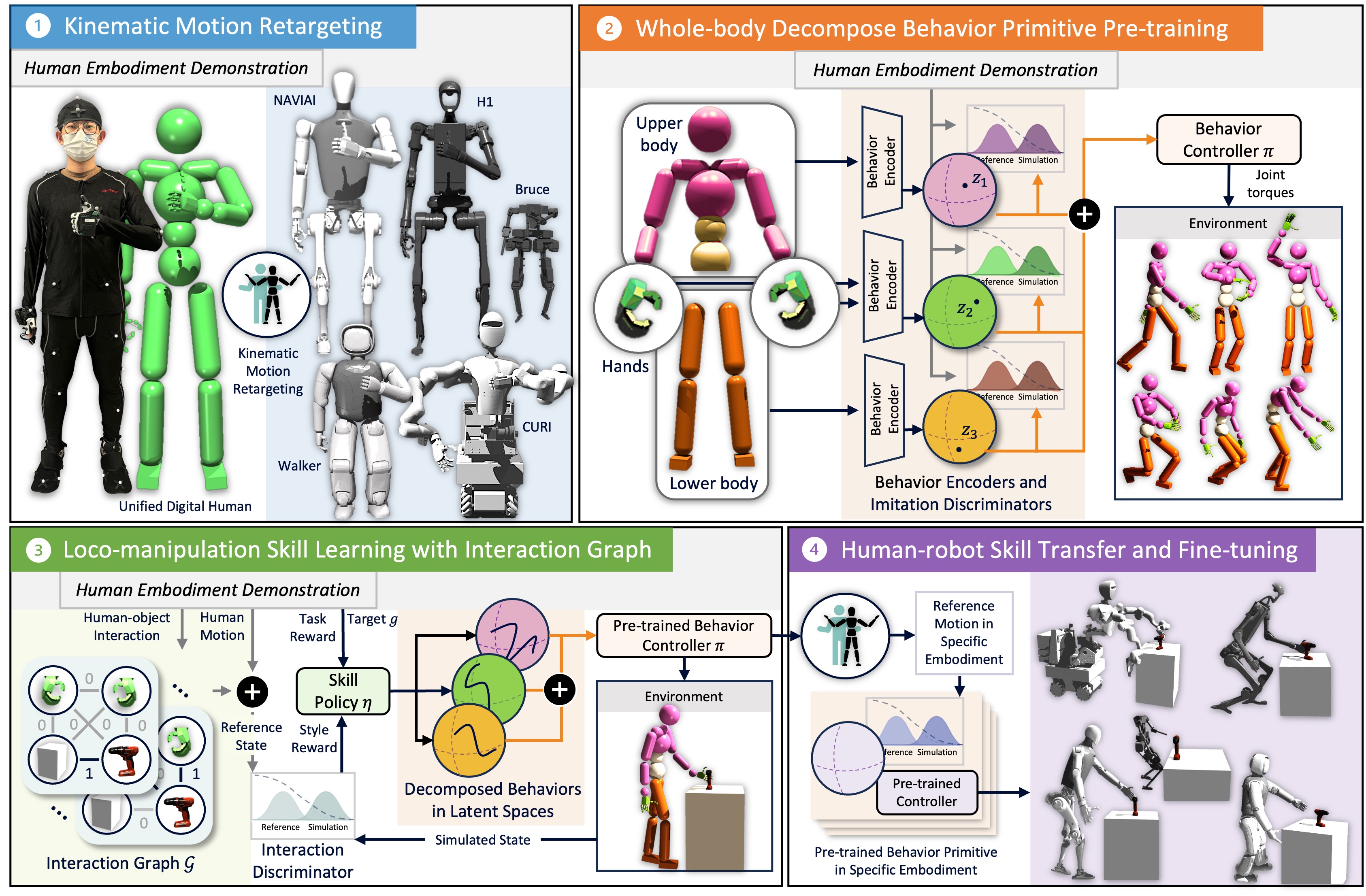}
	\captionsetup{font={small}}
	\caption{\textbf{Schematic overview of the cross-embodiment loco-manipulation skill transfer framework.} 1) Human embodiment demonstration is captured by motion capture system and retargeted to the unified digital human, then retargeted to diverse humanoid robots. 2) High DoFs of humanoids are decomposed into functional parts $p\in [1, P]$ and trained with partial demonstration separately via adversarial imitation to form the latent behavior primitive spaces, humanoids can perform natural and coordinated motions by the behavior controller $\pi(\textbf{a} | \textbf{s}, \textbf{z}_{1\sim P})$.  3) Interaction graphs are extracted from demonstration and guides the policy $\eta( \textbf{z}_{1\sim P}|\textbf{s}, \mathcal{G}, g)$ learning to plan latent behavior trajectories on behavior spaces to complete the interaction skill. 4) By kinematic motion retargeting and adversarial imitation fine-tuning on specific embodiment, same loco-manipulation motions can be deployed on diverse humanoid robots.}
	\label{HOTU}
	\vspace{-1.7em}
\end{figure*}

To address these challenges, we propose a cross-embodiment framework for humanoid behavior and loco-manipulation skill transfer. The proposed framework addresses the configuration differences by introducing a UDH model that serves as a shared prototype to aggregate human demonstrations and abstract embodiment-independent behavior primitives. These primitives encapsulate core motion patterns, which can be retargeted to various humanoid robots through kinematic motion retargeting and embodiment-specific fine-tuning, ensuring compatibility with different robot embodiments. To tackle the challenge of high-dimensional coordination, the framework decomposes the complex control problem into functional components, each trained independently using adversarial imitation learning. Furthermore, a human-object interaction graph is employed to guide the integration of these components, enabling robots to dynamically plan and execute coordinated loco-manipulation tasks. By combining these approaches, the proposed framework achieves robust and generalizable skill transfer across humanoid robots. Extensive experiments on five humanoid robots with varying configurations validate the effectiveness of the proposed framework, which demonstrate its ability to reduce data requirements and generalize learned skills efficiently from human and across different platforms.

\begin{figure*}
	\centering
	\subfigure[]{
		\begin{minipage}[b]{0.3\textwidth}
			\includegraphics[width=1\textwidth]{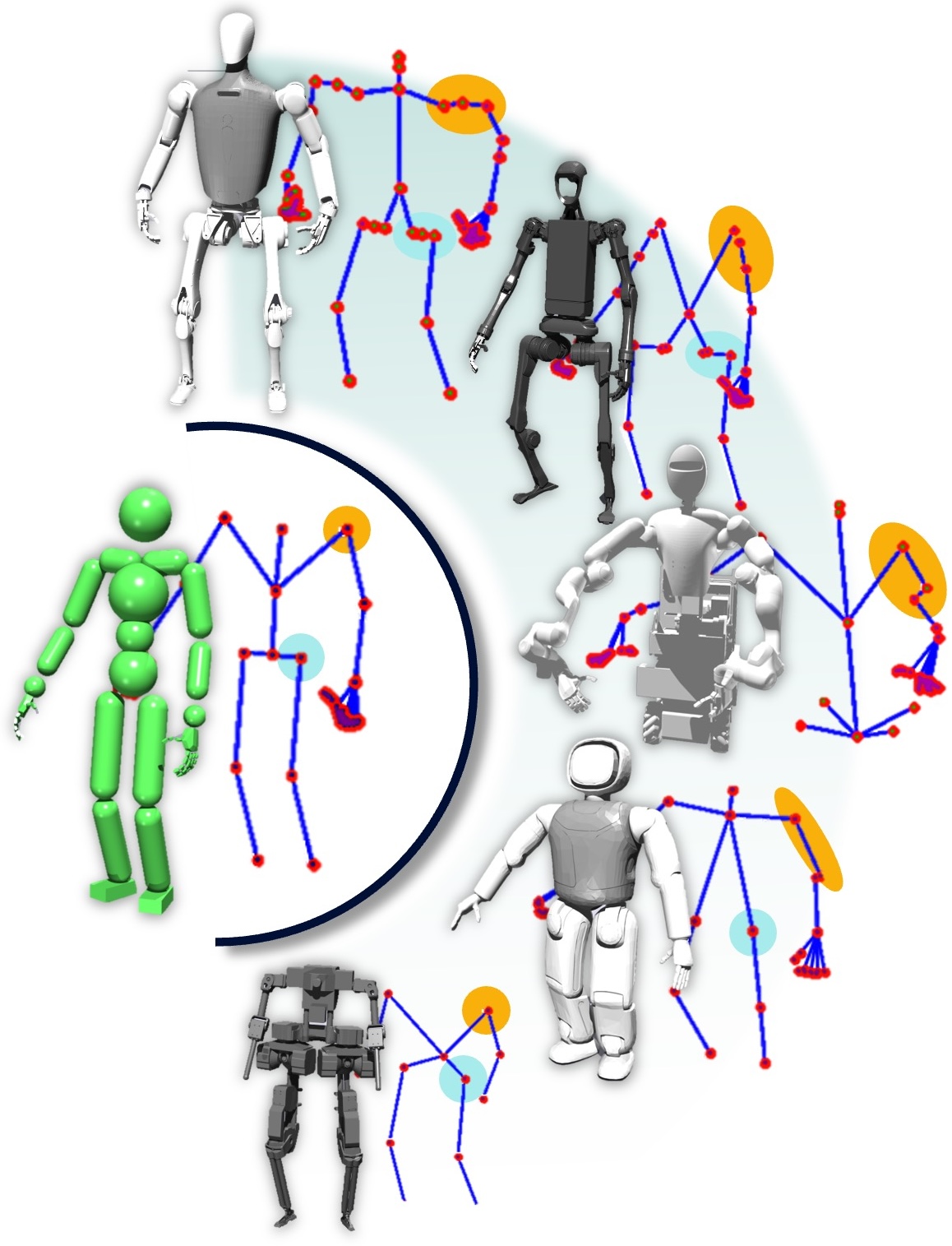}
		\end{minipage}
		\label{motion_retarget}
	}
	\subfigure[]{
		\begin{minipage}[b]{0.51 \textwidth}
			\includegraphics[width=1\textwidth]{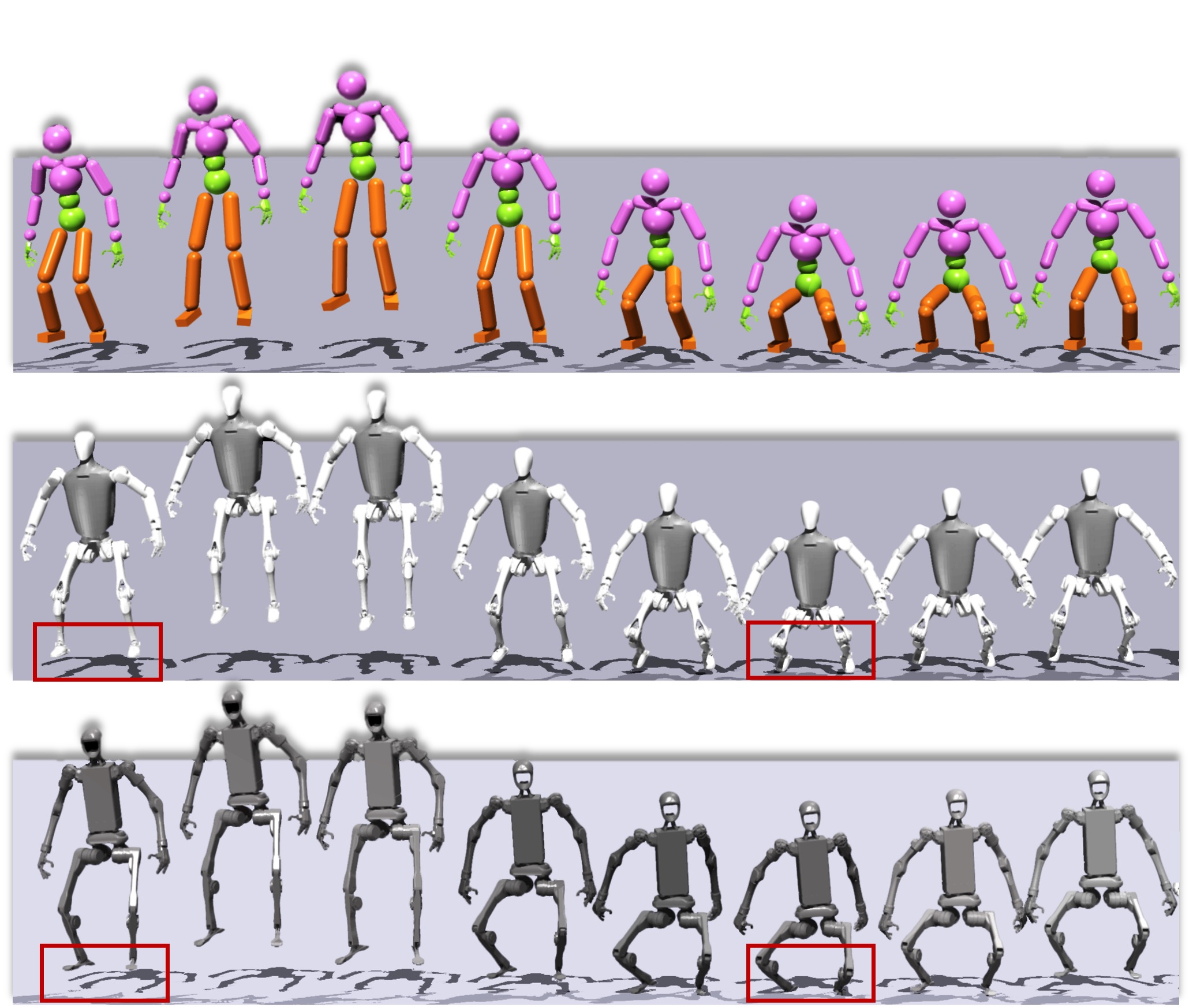}
		\end{minipage}
		\label{motion_retarget_animation}
	}
	\captionsetup{font={small}}
	\caption{\textbf{Kinematic motion retargeting.} (a) Motions are retargeted by grouping DoFs belonging to same function parts and each joint value is solved by partial inverse kinematics. Orange ellipses refer to the group of shoulder DoFs, light blue ellipses refer to the group of hip DoFs. (b) The animation screenshots of kinematic motion retargeting on the unified digital human, NAVIAI and H1 humanoid robots. It is worth noting that the actual dynamics are not considered in this process (see the red boxes), and robots are merely simulating the joint angles, joint velocities, and the trajectories of root links.}
	\label{fig3}
	\vspace{-1.5em}
\end{figure*}

	\section{Related Work}
	Learning-based approaches have become mainstream for enabling humanoid robots to perform locomotion and manipulation tasks. Methods leveraging human demonstrations offer significant benefits by allowing robots to learn directly from human expertise, improving both the efficiency and naturalness of skill acquisition. However, differences in embodiments between humans and robots pose a significant challenge, particularly in transferring skills effectively.
	\vspace{0em}
	
	\subsection{Learning-Based Approaches for Loco-Manipulation}
	Learning-based methods, especially reinforcement learning (RL),  have proven effective for stable locomotion in both quadruped \cite{miki2022learning} and bipedal robots \cite{radosavovic2024real}, but integrating these methods into loco-manipulation introduces challenges. External forces from manipulators can destabilize lower limbs, which recent work addresses by combining learning-based locomotion with model predictive control (MPC) for precise manipulation \cite{ma2022combining}. However, modeling complex tasks like bimanual coordination and dexterous hand manipulation remains difficult, requiring extensive reward shaping \cite{dao2023sim}. 
	
	\subsection{Imitation Learning and Adversarial Imitation Learning}
	Imitation learning is commonly used to overcome the inefficiencies of model-free RL, particularly in humanoids \cite{calinon2009robot}. Traditional methods rely on supervised learning from demonstrations but struggle with generalization \cite{giusti2018flexible,liu2022robot,liu2023softgpt,liu2023birp}. Adversarial imitation learning addresses these limitations by using a discriminator to guide RL exploration towards the distribution of the demonstration data \cite{peng2021amp}. When combined with large human motion datasets, it can enhances high-DoF humanoid skill acquisition and adaptability \cite{peng2022ase}\cite{bae2023pmp}.
	
	\subsection{Cross-Embodiment Skill Transfer}
	Transferring skills across different robotic embodiments has been explored using various approaches. XSkill \cite{xu2023xskill} learns cross-embodiment skill representation from unlabeled videos but is limited to simple manipulation tasks and does not explicitly handle the complexities of humanoid dynamics. Wang et al. \cite{wang2024cross} proposed projecting state and action spaces into a shared latent space for policy transfer across embodiments without task-specific rewards. However, their approach relies on well-defined action mappings, which are challenging to define for humanoid robots due to their high-dimensional and non-linear dynamics. XIRL \cite{zakka2022xirl} learns vision-based reward functions for policy transfer across embodiments. Yang et al. \cite{yang2024pushing} showed that co-training across diverse embodiments improves robustness and enables zero-shot transfer. Mirage \cite{chen2024mirage} uses cross-painting to mask visual differences for successful zero-shot policy transfer. While these works represent significant progress in cross-embodiment skill transfer, they share two key limitations. First, they target simpler table-top robot arms, and do not address the unique challenges of humanoid robots, including high DoF, dynamic balance, and whole-body coordination. Second, they primarily focus on transferring visual representations. However, in humanoid robots, it is essential to additionally account for the embodiment itself, including kinematic and dynamic constrains, which are critical for tasks involving gait dynamics, coordination, and balance.
	

		\begin{table}[t]
		\centering
		\renewcommand{\arraystretch}{1.3}
		\captionsetup{font={small}}
		
		\begin{threeparttable}
			
			\resizebox{0.9\linewidth}{!}{
				\begin{tabular}{c|c|ccccc}
					\Xhline{1px}
					Joints & UDH  &  NAVIAI & H1 & Bruce & Walker & CURI\\
					\Xhline{1px}
					Neck & 3 & 2(YZ) & - & - & 2(YZ) & 2(YZ) \\
					Torso & 3 & 1(Z) & - & - & - & 3 \\
					Shoulder & $3\times 2$ & $3\times 2$ & $3\times 2$ & $2\times 2$(YZ) & $3\times 2$ & $3\times 2$\\
					Elbow & $1\times 2$ & $1\times 2$ & $1\times 2$ & $1\times 2$ & $1\times 2$ & $1\times 2$ \\
					Wrist & $3\times 2$ & $3\times 2$& $3\times 2$& -& $3\times 2$& $3\times 2$\\
					Finger & $1\times 58$ & $1\times 58$ & $1\times 58$ & - & $1\times 20$ & $1\times 58$\\
					Hip & $3\times 2$  & $3\times 2$ & $3\times 2$ & $3\times 2$ & $3\times 2$ & - \\
					Knee & $1\times 2$ & $1\times 2$ & $1\times 2$ & $1\times 2$ & $1\times 2$ & -  \\
					Ankle &  $3\times 2$ &  $2\times 2$(XY) & $1\times 2$(Y) & $1\times 2$(Y)  &  $2\times 2$(XY) & -  \\
					\Xhline{1px}
					Total & 92 & 87 & 83 & 16 & 48 & 77\\
					\Xhline{1px}
				\end{tabular}
			}
		\end{threeparttable}
		\caption{Degrees of freedom for each functional joint part of Unified Digital Human (UDH) and five humanoid robots} 
		\label{DoF}
		\vspace{-2em}
	\end{table}
	
	\section{Cross-embodiment Humanoid \\ Behavior and Skill Transfer}\label{II}
	The schematic overview of the proposed  cross-embodiment humanoid behavior and loco-manipulation skill transfer framework is shown in Fig. \ref{HOTU}. 	
	

		
		\begin{figure*}
			\centering
			\subfigure[]{
				\begin{minipage}[b]{0.62\textwidth}
					\includegraphics[width=1\textwidth]{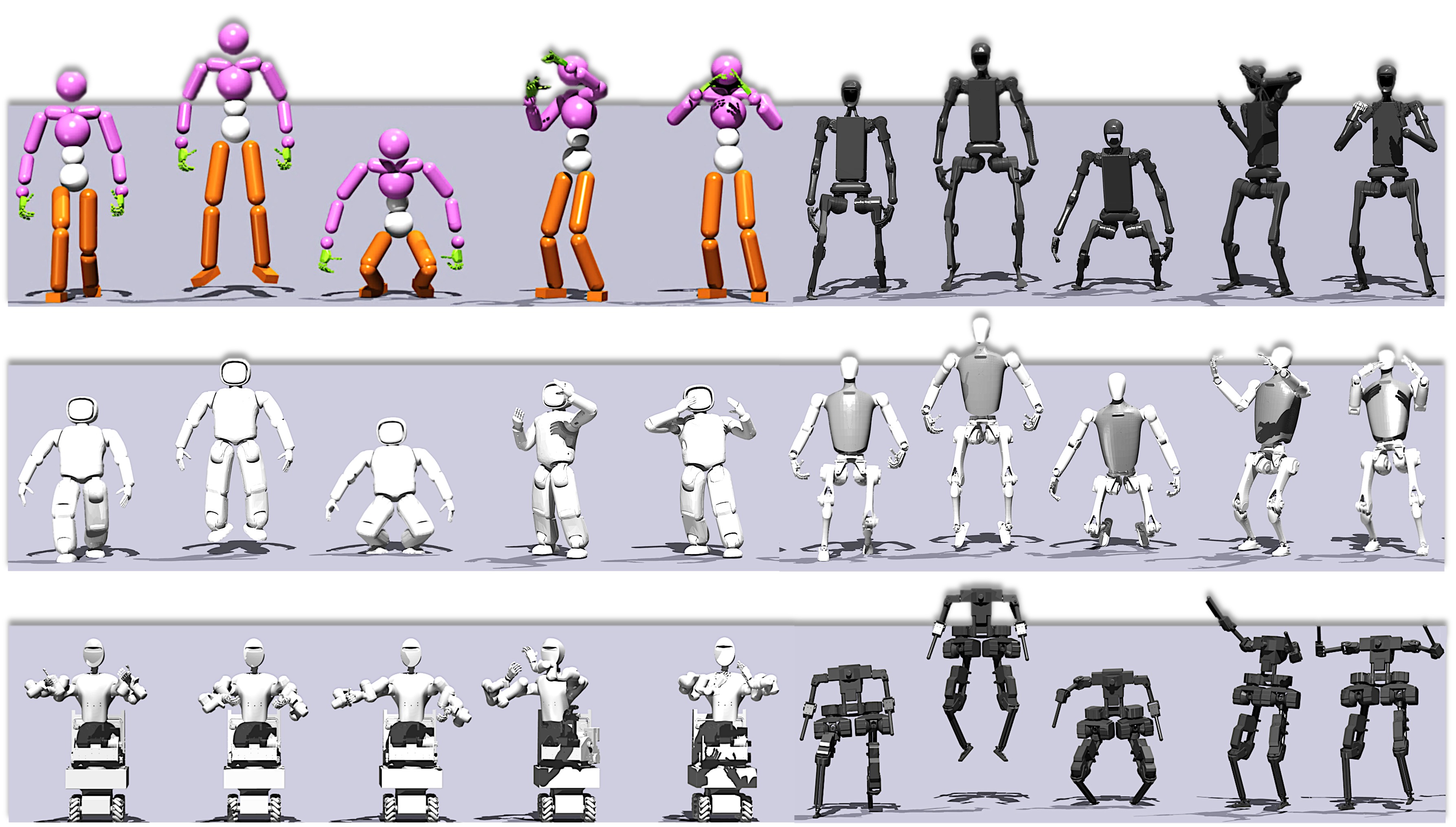}
				\end{minipage}
				\label{behavior}
			}
			\subfigure[]{
				\begin{minipage}[b]{0.24\textwidth}
					\includegraphics[width=1\textwidth]{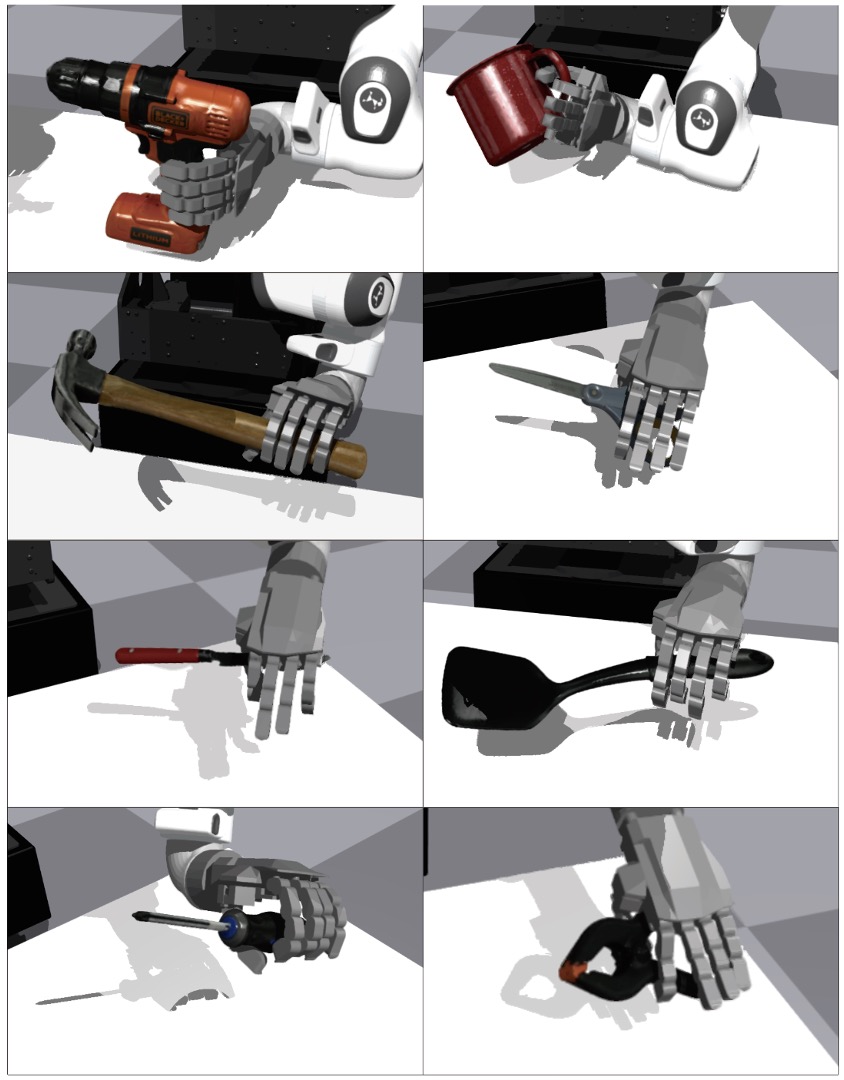}
				\end{minipage}
				\label{hand_pretrain}
			}
			\captionsetup{font={small}}
			\caption{\textbf{Whole-body functional decomposition and behavior primitive training.} (a) Behavior primitive pre-training for the unified digital human and five humanoid robots. It shows the style imitation of the given motion dataset using pre-trained behavior primitives. (b) Hand-specific behavior primitive pre-training.}
			\label{fig4}
			\vspace{-1.5em}
		\end{figure*}

		\subsection{Kinematic motion retargeting}\label{IIA}
		Humans possess complex joint structures, with many joints capable of 3-DoF coaxial rotation. Previous work has utilized the $51\times3$-DoF SMPL human model \cite{SMPL:2015}, which represents all human joints as 3-DoF spherical joints. However, joints like elbows, knees, fingers, and ankles can be simplified to 1- or 2-DoF. This simplification reduces computational complexity and better aligns with the joint structures used in humanoid robots, which typically replicate higher DoF joints using multiple motors. Consequently, a unified digital human model with 92 DoFs (Table \ref{DoF}) was designed. Our framework was tested on five humanoid robots with diverse configurations: NAVIAI from Zhejiang Humanoid Robot Innovation Center, H1 from Unitree, Walker from Ubtech, Bruce from Westwood Robotics, and CURI from the CLOVER lab of The Chinese University of Hong Kong. 
		
		To enable effective cross-embodiment transfer, we use a kinematics-based motion retargeting technique, grouping multiple DoFs by joint functionality (Table \ref{DoF}). For example, Fig. \ref{motion_retarget} shows the functional groups for the shoulder (orange) and hip (blue) across all robots. Within each group, multiple DoFs are treated as a simplified open-chain manipulator. DoF positions and velocities are calculated using partial inverse kinematics for efficient, accurate retargeting. Root link trajectories, like the pelvis, are normalized to account for the differences in scale between the human demonstrator and humanoid robots. This normalization addresses mismatches in stride length and actual displacement caused by size discrepancies, ensuring that the retargeted trajectories align appropriately with the physical dimensions of robots. Fig. \ref{motion_retarget_animation} shows example retargeting results. It is important to note that this process focuses on kinematics (see red boxes), dynamics like joint torques and gravity will be considered in following sections. The retargeted motions generated serve as reference motions for subsequent behavior primitive pre-training.
		\vspace{-0.5em}

		\subsection{Whole-body Functional Decomposition and Behavior Primitive Training}\label{IIB}
		
		Demonstration datasets typically cover only a limited range of motions and cannot encompass all possible behaviors. Therefore, the goal of imitation learning is to capture the behavioral characteristics rather than directly replicating the motions in the dataset. These behavioral characteristics, known as behavior primitives, are fundamental building blocks of motion that represent basic patterns of demonstrations. By combining these primitives, humanoid robots can adapt to diverse tasks and generate complex behaviors.
		
		To achieve this, a low-level behavior controller $\pi(\textbf{a} | \textbf{s}, \textbf{z})$ is introduced, which generates actions $\textbf{a}$ based on the current state $\textbf{s}$ and a latent behavior variable $\textbf{z} \in \mathcal{Z}$, sampled from a prior distribution $p(\textbf{z})$. The behavior latent space $\mathcal{Z}$, defined as a unit hypersphere, represents feasible humanoid behaviors by capturing generalizable patterns from human demonstrations. This design decouples high-level task planning from low-level motion execution, enabling flexible and reusable behavior generation.
		
		To train the low-level controller, adversarial imitation learning is employed, where the RL policy acts as a generator, and a discriminator $D$ is designed to distinguish real human motions from generated behaviors. This approach ensures that the controller learns to produce actions that align with the behavioral characteristics of the unstructured motion dataset $\mathcal{M}_u$, rather than merely replicating specific trajectories. The objective of the discriminator is defined as:
		\begin{equation}
			\begin{small}
				\begin{aligned}
					\min _{D} & -\mathbb{E}_{d^{\mathcal{M}_u}\left(\textbf{s}, \textbf{s}^{\prime}\right)}\left[\log \left(D\left(\textbf{s}, \textbf{s}^{\prime}\right)\right)\right]-\mathbb{E}_{d^{\pi}\left(\textbf{s}, \textbf{s}^{\prime}\right)}\left[\log \left(1-D\left(\textbf{s}, \textbf{s}^{\prime}\right)\right)\right] 
			\end{aligned}\end{small}\label{adversarial}
		\end{equation}
		
		In this setting, the \textbf{behavior style reward} is derived from the output of the style discriminator: $r_b\left(\textbf{s}, \textbf{s}'\right)=\max [0,1-0.25\left(D\left(\textbf{s}, \textbf{s}'\right)-1\right)^2]$, which measures the similarity between the generated motions and real human motions. The final pre-trained policy $\pi$ is expected to generate natural and realistic humanoid behaviors for any given latent skill $\textbf{z}$. Since a reference motion dataset is available for each robot through kinematic motion retargeting, embodiment-specific behaviors can be pre-trained using this same approach.
		
		\begin{figure*}[t]
			\centering
			\includegraphics[width=0.65\linewidth]{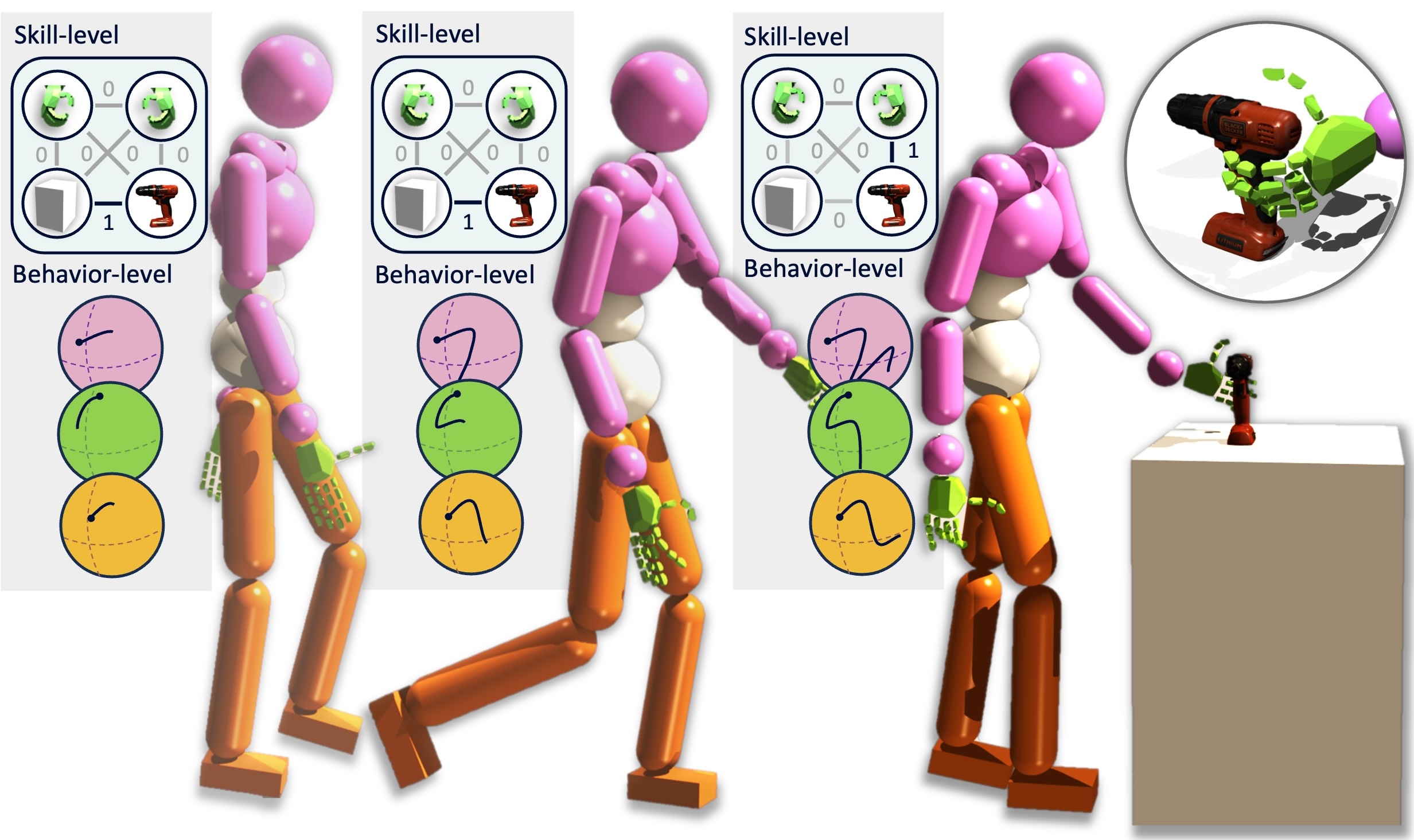}
			\captionsetup{font={small}}
			\caption{\textbf{Loco-manipulation skill learning with interaction graph.} An illustration of the detailed loco-manipulation skill learning process with skill-level interaction graph guidance and decomposed behavior primitive trajectories in latent behavior spaces.}
			\label{loco_manipulation_process}
			\vspace{-1.7em}
		\end{figure*}
		
		To address the challenges posed by the high redundancy and coordination requirements of humanoid robots, we propose a method called \textbf{decomposed adversarial imitation learning (DAIL)}, as illustrated in Part 2 of Fig.~\ref{HOTU}. In this approach, the entire humanoid body is decomposed into several functional parts, such as lower limbs for walking, upper limbs for manipulation, and hands for interacting with objects. These components are trained using whole-body motion data but can also be targeted using partial datasets. This decomposed approach is particularly useful for handling different parts that require varying levels of control precision, such as dexterous hands that necessitate fine manipulation versus limbs that rely on gross movements.  The decomposed module includes a behavior encoder and a style discriminator for each functional part. During the whole-body behavior controller training, style rewards and losses from each part are multiplied together. This enables effective learning and coordination of behaviors across different functional parts. 
		
		After training behavior primitives via DAIL, the humanoid model can perform human-like motions while considering dynamics and joint limitations, compared to pure kinematic retargeting as discussed in Sec. \ref{IIA}. The results are shown in Fig. \ref{fig4}, and the detailed process for decomposed pre-training is elaborated in Sec. \ref{IIIC}.

		\subsection{Loco-manipulation skill learning with interaction graph}\label{IIC}
		
		\textbf{Loco-manipulation skills can be constructed based on previous locomotion and whole-body free motion behaviors, guided by human-object interaction in an object-centric manner.} In this section, the concept of an interaction graph $\mathcal{G}$ is introduced to describe human-object interactions. Once the low-level controller $\pi$ with pre-trained human behavior primitives is available, a goal-oriented high-level policy $\eta(\textbf{z}|\textbf{s}, \mathcal{G}, g)$ can be designed to generate a sequence of latent behaviors $\textbf{z}$ that accomplish the task.
		
		The motivation of interaction graph is to handle the variability of real-world interactions. Although motions of interacted objects are captured during human embodiment demonstrations, directly using these object trajectories for imitation learning is not appropriate. Robots are likely to encounter various task parameters with same interaction skills, like different start and end poses or object sizes. Among them, the interaction pattern between humans and objects remains relatively consistent. Thus, in this work, an interaction graph is extracted and constructed for each time step from demonstration to represent interaction skills.
		
		
	The interaction graph $\mathcal{G}$ is a structured representation where nodes represent body parts and objects, and edges indicate interactions such as contact or proximity. Each node encodes the pose (position and orientation) of the corresponding body part or object, while edges are binary indicators of contact: $1$ for contact, $0$ otherwise. To handle the variability of task parameters, the relative distances between the hands and the object are encoded in the hand node. Similarly, the relative distance between the current pose of object and its target pose is stored in the object node. In loco-manipulation tasks, the focus is primarily on hand-object contact. To reduce complexity, body parts like hands, composed of multiple rigid bodies, are aggregated into a single node. This aggregation keeps the graph manageable, minimizing noise and avoiding excessive size.

		
		
		To ensure that the interaction style of robot aligns with the human demonstration, a style discriminator $D_I$ is introduced. This discriminator evaluates the similarity of the robot interaction graph to the reference graph from human demonstrations. The state input for the interaction style discriminator $D_I$ is constructed as $\textbf{s}_I = \{d_{ho}, d_{og}, e_{ho}\}$, where $h$ refers to hands, $o$ refers to the object, and $g$ refers to the target. $d_{ho}$ and $d_{og}$ represent the relative distance information, and $e_{ho}$ denotes the binary contact value between the hands and the object. The specific loss function for $D_I$ is then given as:
		\begin{equation}
			\begin{small}
				\begin{aligned}
					\min _{D_I}  -\mathbb{E}_{d^{\mathcal{M}}\left(\textbf{s}_I, \textbf{s}^{\prime}_I\right)}&\left[\log \left(D_I\left(\textbf{s}_I, \textbf{s}^{\prime}_I\right)\right)\right]\\
					&-\mathbb{E}_{d^{\eta}\left(\textbf{s}_I, \textbf{s}^{\prime}_I\right)}\left[\log \left(1-D_I\left(\textbf{s}_I, \textbf{s}^{\prime}_I\right)\right)\right] 
			\end{aligned}\end{small}\label{adversarial2}
		\end{equation}
		where $\mathcal{M}$ is the human embodiment demonstration dataset with human-object interactions already retargeted to the unified digital human model. For training the high-level skill policy $\eta$, the reward function is a combination of task reward and style reward. The style reward contains both a high-level interaction skill style $s$ and a low-level behavior style $b$:
		\begin{equation}
			\begin{small}
				\begin{aligned}
					r=w_g r_g\left(\textbf{s}, \textbf{a}, \textbf{s}^{\prime}, g\right) &+ w_{s} r_s(\textbf{s}_I, \textbf{s}^{\prime}_I)+w_{b}r_b(\textbf{s}, \textbf{s}^{\prime})
				\end{aligned}
			\end{small}
			\label{skill_reward}
		\end{equation}
		where $w_*$ are the weights for each reward component and \textbf{skill style reward}  $r_s(\textbf{s}_I, \textbf{s}^{\prime}_I)=\max [0,1-0.25\left(D_I\left(\textbf{s}_I, \textbf{s}'_I\right)-1\right)^2]$.
		
		An example of the complete loco-manipulation process using the proposed framework is shown in Fig.~\ref{loco_manipulation_process}. At the skill level, the humanoid imitates the interaction pattern to achieve proper hand-object interaction, generating trajectories in latent behavior spaces that produce the actual motion at the behavior level.

		\subsection{Human-Robot Skill Transfer and Fine-Tuning}\label{IID}

		Leveraging the pre-training of behavior primitives across diverse humanoid embodiments and the generalizable skill learning on a unified digital human, the deployment of consistent skills on different humanoid robots is significantly streamlined. The methodology outlined in this section (Part 4 in Fig.~\ref{HOTU}) builds upon the earlier sections.
		
		The foundation model for loco-manipulation skills is established through pre-trained skill policies on the unified digital human. Task execution motions for each specific humanoid robot embodiment are retargeted from this foundation model, eliminating the need for re-training and enabling efficient skill deployment. The retargeting process involves executing loco-manipulation tasks based on the unified digital human embodiment and subsequently mapping the generated motion to the desired humanoid robot embodiment using kinematic motion retargeting techniques, as described in Sec.~\ref{IIA}. With the retargeted motion as a reference, a similar high-level control structure can be constructed as discussed in Sec.~\ref{IIC}.
		
		It is important to note that the skill policy used during the fine-tuning phase differs from and is simpler than the one described in Sec.~\ref{IIC}. In the previous section, the reference trajectory consisted of a small amount of demonstration data with varying task parameters, making the training process more challenging. However, during fine-tuning, the objective is to directly imitate the retargeted trajectory that is already capable of achieving the desired task. 
		
		To facilitate this, a simple Multi-Layer Perceptron (MLP)-based fine-tuning layer is introduced. This layer generates hyper curves in specific behavior primitive spaces, converting the kinematics-level retargeted motion into dynamics-level motion that produces actual control commands. The fine-tuning process is crucial for adapting the retargeted motion to the dynamics and control requirements of the specific humanoid robot embodiment, ensuring that the robot can further perform the task effectively in real-world environments.

		\section{Experiments}\label{III}
		
		\subsection{Setup}\label{IIIA}
		
		\textbf{State:} The state representation of UDH and each humanoid robot includes the position and velocity of every joint, as well as the translation, rotation, velocity, and angular velocity of the root link. Additionally, the global poses of key body parts, such as the hands during object manipulation tasks, are included. The interaction state, which comprises object-related information, has been detailed in Sec.~\ref{IIC}.
		
		\textbf{Action:} The behavior controller outputs target joint positions, which are executed by a Proportional-Derivative (PD) controller that generates the necessary joint torques. The dimensions of the joints for each humanoid robot are listed in Table~\ref{DoF}. It should be noted that, \textbf{motor torque limitations were simulated by constraining the output of PD controller}, though the resulting dynamics may differ from real-world conditions. Joint limitations and inertia for each humanoid were defined based on their URDF models which are similar as hardware constraints.
		
		\textbf{Performance Metrics:} Behavior primitives are not task-oriented and lack clear task completion metrics. Therefore, the training return for the behavior primitives is fully represented by the style reward provided by the discriminator. In the skill learning stage, the training return is given by Equ. \ref{skill_reward}, reflecting both task completion and whether the robot behavior is anthropomorphic.
		
		\textbf{Skill Policy and Behavior Controller:} The policy networks employed to control the humanoid robots are based on an actor-critic framework and utilize the Proximal Policy Optimization (PPO) algorithm, a widely used method in RL.
		
		\textbf{Humanoid Models and Simulation Environment:} The five humanoid robot models used in this study-NAVIAI, H1, Bruce, Walker and CURI—underwent refinement to ensure accurate simulation, including self-collision checks and the addition of extra hands for the NAVIAI and H1 robots. These models are publicly available on the Rofunc platform \cite{liu2023rofunc}. The experiments were conducted using Isaac Gym \cite{makoviychuk2021isaac}, a high-performance, GPU-based physics simulator designed specifically for robot learning applications. To enhance the robustness of the humanoid models, those equipped with legs (NAVIAI, H1, Bruce, and Walker) were initialized in each simulation episode by dropping from mid-air, prompting them to stabilize and maintain an upright stance—a practice that helps train the models for dynamic balance.

		\begin{figure*}[t]
			\centering
			\includegraphics[width=0.9\linewidth]{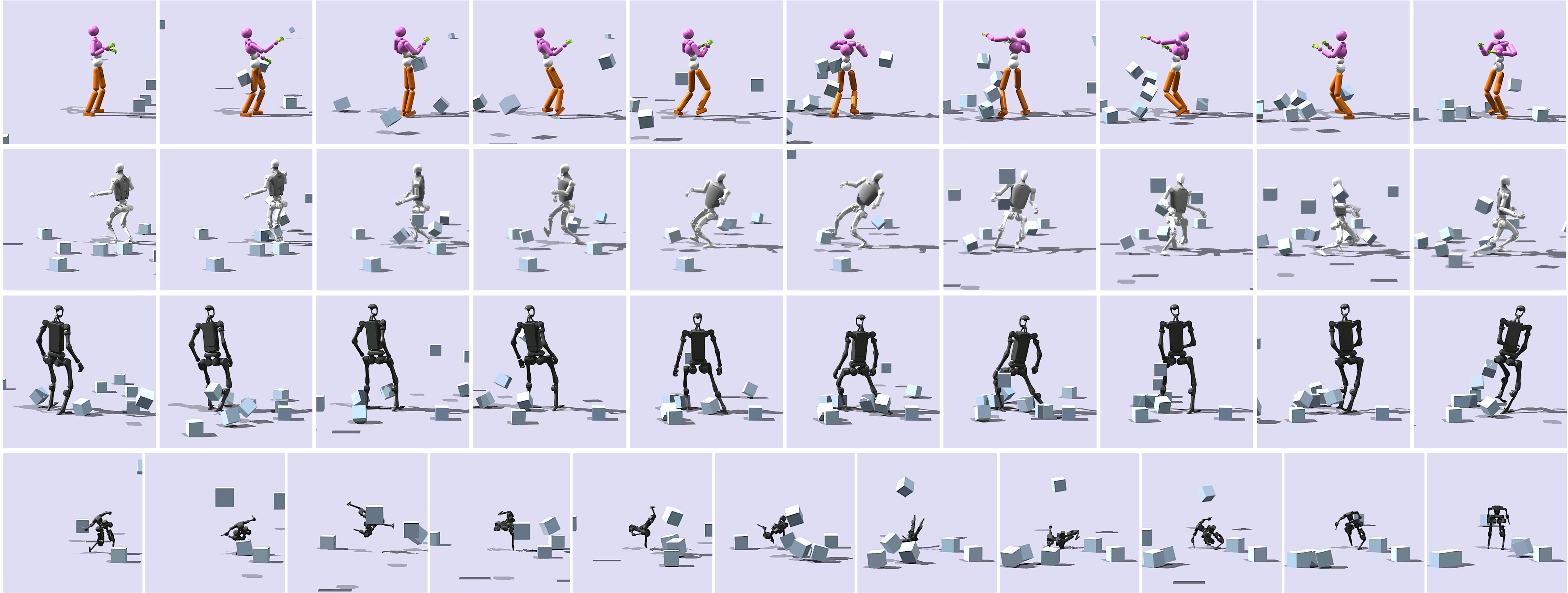}
			\captionsetup{font={small}}
			\caption{\textbf{Behavior primitive pre-training.} Stability experiments for legged humanoid robots by hitting them with moving cubes.}
			\label{perturb}
			\vspace{-1em}
		\end{figure*}

		\subsection{Data Collection and Motion Dataset}\label{IIIB}
		
		The human embodiment demonstrations were captured using the Optitrack motion capture system, as illustrated on the left side of Fig.~\ref{cover}. To accurately capture whole-body motions, a motion capture suit equipped with 41 markers was utilized. Hand movements were recorded using Manus gloves, which provide detailed tracking of finger and hand motions. Additionally, markers were placed on objects involved in the tasks to track their poses and interactions with the human subject.
		
		The collected data, which includes comprehensive recordings of human motions, were converted to the FBX format using Optitrack software. These FBX motion files were subsequently parsed and retargeted to various humanoid robot embodiments using the Rofunc platform. Rofunc automates the retargeting process, ensuring that the motion data is appropriately adapted to different robotic configurations.
		
		The unstructured human motion dataset we gathered encompasses a wide range of actions, reflecting diverse everyday and specialized movements. These include walking, jumping, squatting, boxing, finger movements, handshakes, Kung Fu, leg stretching, waving, slapping, chest pressing, and giving a thumbs up. Additionally, two specific loco-manipulation tasks are included: walking while carrying a box and walking to grasp a power drill. These tasks were selected to demonstrate the framework capability in transferring complex, coordinated behaviors involving both locomotion and manipulation.

				\begin{table}[t]
			\centering
			\renewcommand{\arraystretch}{1.3}
			\captionsetup{font={small}}
			
			\begin{threeparttable}
				
				\resizebox{0.85\linewidth}{!}{
					\begin{tabular}{c|c|ccccc}
						\Xhline{1px} 
						Tasks 			 & UDH &  NAVIAI  & H1 		& Bruce 	& Walker & CURI		\\
						\Xhline{1px}
						Walk  			   & 0.93  & 0.81       & 0.82 	& \textbf{0.95}		 & 0.87 	& -				\\
						Jump 			  & 0.91  & 0.77 	   & 0.79  & 0.91 		& 0.73   	& -				\\
						Squat 			  & 0.94 & 0.82 	  & 0.89   & \textbf{0.96}      & 0.86 & -					\\
						Leg stretch   & 0.93 & 0.84 	  & 0.90 & \textbf{0.94}        & 0.87 & -					\\
						Box 			    & 0.93 & 0.91    	& 0.93 & \textbf{0.98}           & 0.90 & 0.83			\\
						Finger 			   & 0.83 & 0.80 	  & 0.83 & -                 & 0.76 & 0.82				\\
						Handshakes & 0.90 & 0.87 	   & 0.93 & \textbf{0.95}         & 0.84 & 0.88				\\
						Kung Fu 		& 0.91 & 0.86        & 0.91 & 0.91         & 0.74 & 0.85			\\
						Wave 			 & \textbf{0.95}& 0.92 		  & 0.90 & 0.91         & 0.87 & 0.91			\\
						Slap 			    & \textbf{0.95} & 0.92 		& 0.93 & 0.91         & 0.87 & 0.90				\\
						Thump up 	 & \textbf{0.92} & 0.93 		 & 0.92 & -               & 0.90 & 0.87			\\
						\Xhline{1px}
						Hand grasp   & -        & - 		     & - 		& -               & -       & 0.73\\
						\Xhline{1px}
					\end{tabular}
				}
			\end{threeparttable}
			\caption{\textbf{Behavior primitive pre-training.} Normalized average training returns of UDH and five humanoid robots for each tasks.} 
			\label{table2}	
			\vspace{-1.7em}
		\end{table}
		
		To support the reproducibility of our work and to facilitate further research, we have made the retargeted motion dataset available for the five humanoid robots used in our experiments, as well as the unified digital human model. These datasets can be accessed through the Rofunc website.

		\subsection{Decomposed Behavior Primitive Pre-Training}\label{IIIC}
								\begin{figure}[t]
			\centering
			\includegraphics[width=0.9\linewidth]{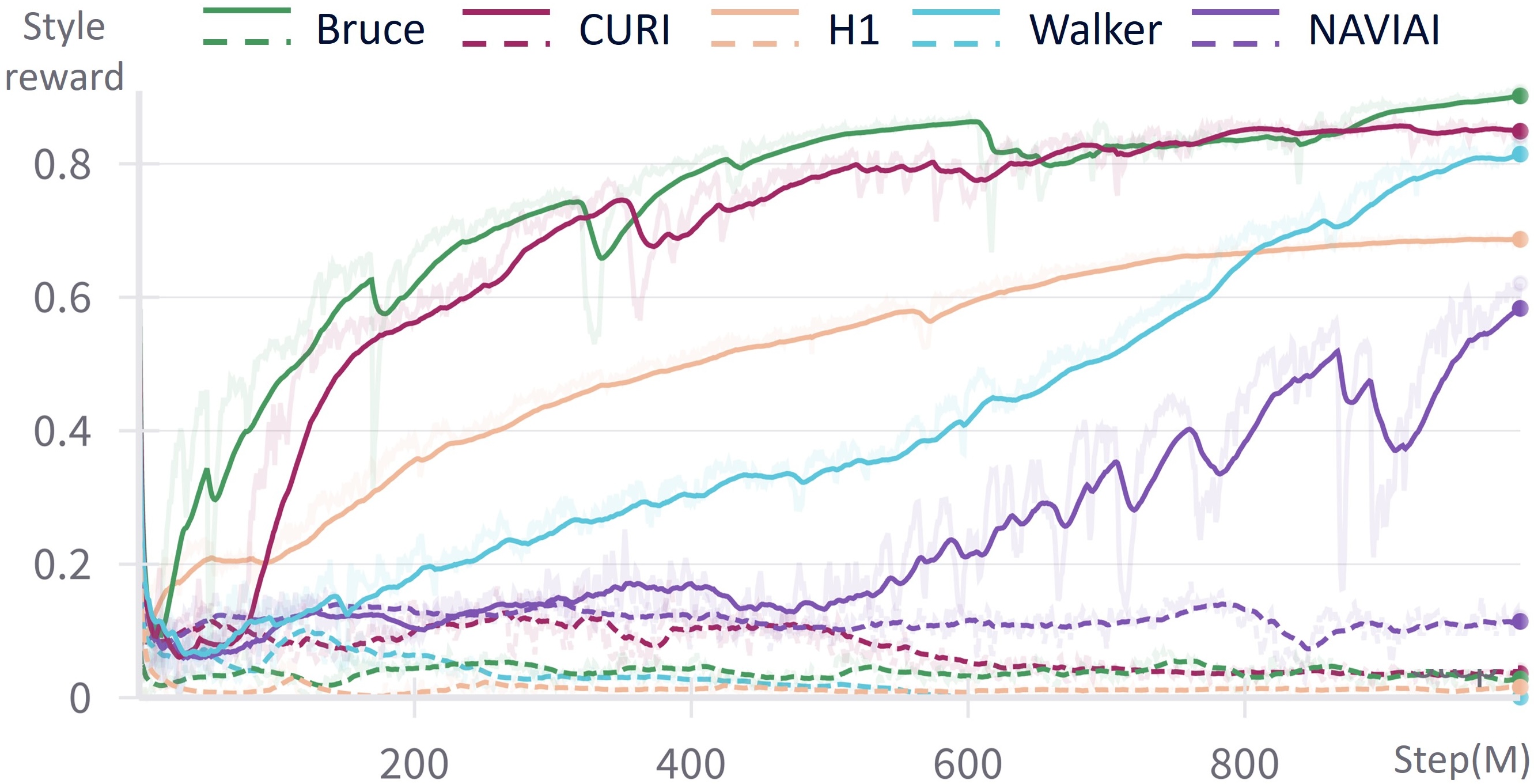}
			\captionsetup{font={small}}
			\caption{\textbf{Behavior primitive pre-training.} Ablation study for the effectiveness of decomposition.  Solid lines refer to decomposed imitation learning, dashed lines refer to the learning without decomposed imitation.}
			\label{reward}
			\vspace{-1.8em}
		\end{figure}
		Decomposed behavior primitive pre-training is conducted independently on the UDH and five humanoid robots. The reference motion dataset for training was transferred from the human demonstrator to each robot embodiment via kinematic motion retargeting. After pre-training with the entire motion dataset, the imitation of specific behaviors can be fine-tuned using the corresponding reference motion data. Formally, this process identifies the latent subspace associated with the specific behavior within the overall behavior latent space. The promising results of imitating these behaviors are illustrated in Fig.~\ref{behavior}, which showcases the style imitation capabilities achieved by the pre-trained behavior primitives when applied to the specific motion data. The detailed performance of each humanoid across all tasks is presented in Table \ref{table2}. The normalized average training returns reflect both motion similarity to the reference and locomotion stability. Notably, Bruce performed exceptionally well in most tasks due to its relatively lower DoFs. CURI, equipped with a wheeled base, showed strong performance in locomotion. However, its overall performance was limited by the configuration differences between its Franka arms and typical humanoid arms. The three full-sized humanoid robots (NAVIAI, HI, and Walker) have the highest number of DoFs and more complex structures compared to the UDH, which led to slightly lower overall performance.

		One of the most challenging aspects of training decomposed behavior primitives is the high degree of freedom associated with hand motions. To address this challenge, we conducted \textbf{specific pre-training focused on the hand component}. This was achieved using a grasping dataset, which was collected with the qbSofthand and the YCB dataset. The dataset enabled us to train finger grasping motions specifically for the qbSofthand, which is equipped on the CURI robot, as shown in Fig.~\ref{hand_pretrain}. After pre-training, the behavior primitive for hand motion was integrated into the whole-body framework and further refined alongside other body parts using the unstructured human motion dataset, as described in Sec.~\ref{IIB}.
		
		To further evaluate the stability and robustness of the legged humanoid robots, we designed a perturbation experiment where moving cubes were directed at the robots to simulate external disturbances. The outcomes of this experiment are depicted in Fig.~\ref{perturb}, showing how the different robots responded to these perturbations. The results demonstrate that the pre-trained behavior primitives enable the robots to maintain stability under dynamic conditions. For a more detailed analysis, corresponding videos demonstrating these experiments are available on the Rofunc website\footnote{\url{https://rofunc.readthedocs.io/en/latest/lfd/RofuncRL/HOTU.html}}.
				
		An ablation study was conducted to demonstrate the effectiveness of decomposition in skill imitation learning. The style reward curves for the five humanoid robots, shown in Fig.~\ref{reward}, clearly highlight the impact of DAIL. A key finding is the comparison between the imitation performance of humanoid robots with (solid lines) and without (dashed lines) whole-body decomposition. The decomposed imitation method consistently outperforms the non-decomposed approach, demonstrating the significant advantages of whole-body decomposition in accelerating the learning process and enhancing performance, particularly for high DoFs robots.

		\begin{figure*}[t]
	\centering
	\includegraphics[width=0.83\linewidth]{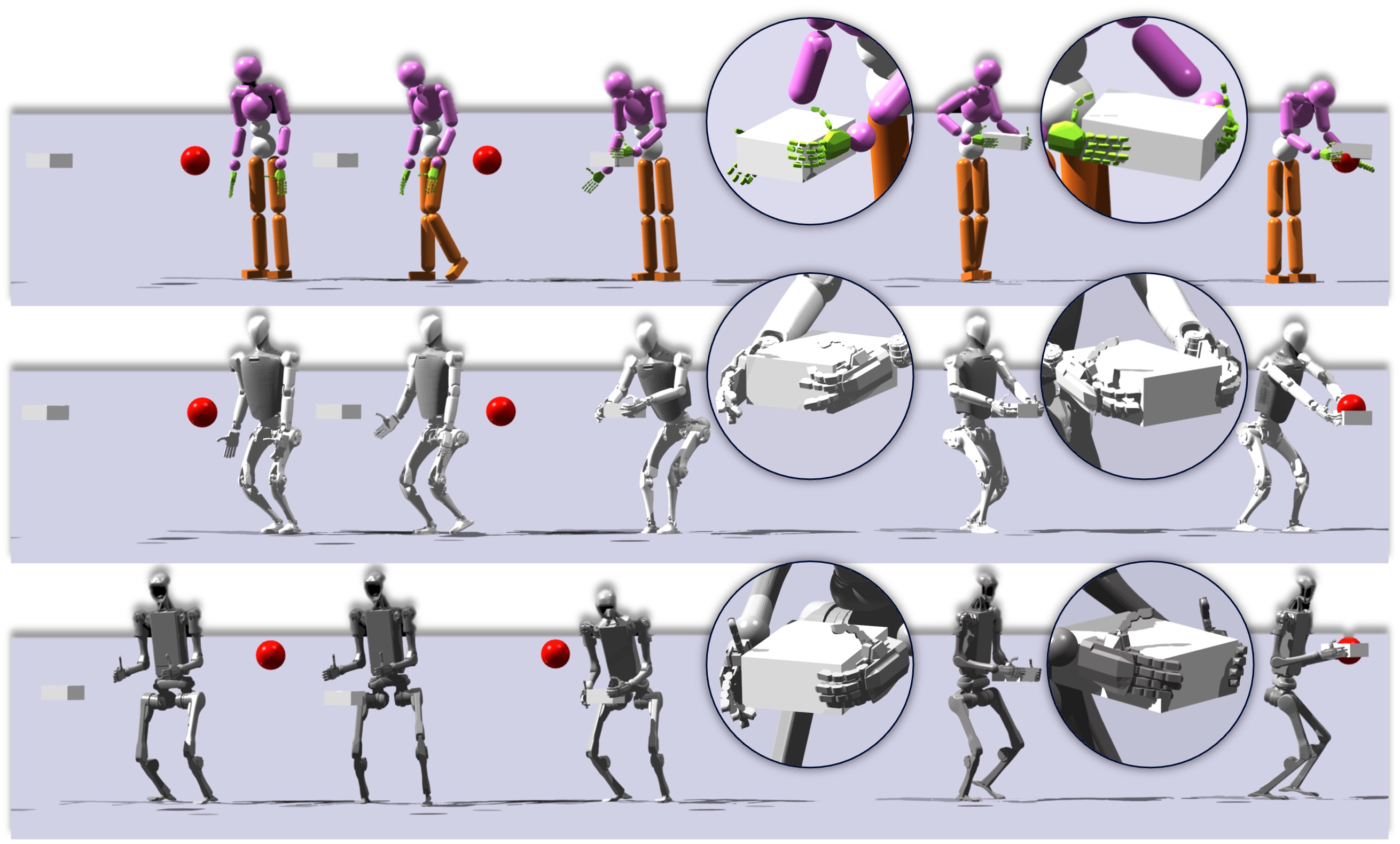}
	\captionsetup{font={small}}
	\caption{\textbf{Loco-manipulation skill.} Box carrying and putting task performed by unified digital human, NAVIAI and UnitreeH1 robots.  For better visualization, the box is set to float at the start position, and the red sphere is the target position of the box. }
	\label{final_exp}
	\vspace{-1em}
\end{figure*}

		\begin{table}[t]
			\centering
			\renewcommand{\arraystretch}{1.3}
			\captionsetup{font={small}}
			\begin{threeparttable}
				
				\resizebox{1.0\linewidth}{!}{
					\begin{tabular}{c|c|ccccc}
						\Xhline{1px}
						Methods & UDH &  NAVIAI & H1 & Bruce$^*$ & Walker & CURI\\
						\Xhline{1px}
						AMP & 0.32/2.36h & 0.21/2.60h & 0.25/2.57h & - & 0.24/2.89h & 0.40/2.21h\\
						ASE  & 0.48/4.53h & 0.37/5.23h & 0.53/4.89h & - & 0.46/4.77h & 0.68/4.34h\\
						PMP  & \textbf{0.88}/4.96h & 0.77/5.81h & \textbf{0.83}/5.44h & - & 0.80/5.62h & 0.79/4.79h\\
						\Xhline{1px}
						Ours & 0.87/\textbf{4.37h} & \textbf{0.84}/\textbf{0.97h} & 0.81/\textbf{0.92h} & - & \textbf{0.85}/\textbf{0.87h}& \textbf{0.89}/\textbf{0.75h}\\
						\Xhline{1px}
					\end{tabular}
				}
				\begin{tablenotes}
					\footnotesize
					\item[$ * $] Bruce was excluded due to the lack of hands.
				\end{tablenotes}
			\end{threeparttable}
			\caption{\textbf{Loco-manipulation skill.} Comparison of normalized average returns and training hours of UDH and five humanoid robots for box carrying task.} 
			\label{table3}
			\vspace{-2em}
		\end{table}
		
		\subsection{Loco-Manipulation Skill and Fine-Tune Performance}\label{IIID}
		
		Figure~\ref{final_exp} illustrates a loco-manipulation task where humanoid robots walk towards a floating box, grasp it, and place it at a designated target position. The first two rows depict the same task but are executed by the UDH and NAVIAI, respectively. This demonstrates the process of motion retargeting and fine-tuning, showcasing how a skill learned from human demonstrations can be effectively transferred to different humanoid robot embodiments. The third row of Fig.~\ref{final_exp} features the UnitreeH1 robot performing the same loco-manipulation task, but with different starting and target positions for the box. This variation highlights the generalization capability of the learned loco-manipulation skill, showing that the skill is not merely overfitted to specific scenarios but can adapt to new task parameters across different robot embodiments.
		
		To evaluate the proposed framework, performance of box carrying skill learning and the training time required has been compared against following existing methods.
			
			\begin{itemize}
				\item \textbf{AMP}  \cite{peng2021amp}: AMP can be viewed as a baseline that only includes the behavior primitive pre-training stage.
				\item \textbf{ASE}  \cite{peng2022ase}: ASE is a hierarchical framework that builds upon AMP by introducing skill learning but without the interaction graph guidance. Comparing with ASE highlights the importance of the interaction graph.
				\item \textbf{PMP}  \cite{bae2023pmp}: PMP applies the decomposition concept to a similar UDH. Comparing with PMP demonstrates the effectiveness of our method in improving training efficiency by cross-embodiment transfer.
			\end{itemize}
			
			As shown in Table \ref{table3}, our framework significantly improves task completion rates and reduces training time. The results demonstrate that our framework, which combines kinematic motion retargeting, interaction graph guidance, and embodiment-specific fine-tuning, is more effective for loco-manipulation skill transfer across humanoid robots.

		\section{Conclusion}\label{V}
		
		In this article, we presented a framework for cross-embodiment loco-manipulation skill transfer from human demonstrations to various humanoid robots. This approach addresses the challenges of configuration differences and the high degrees of freedom in diverse humanoid embodiments. The framework leverages a unified digital human model as a common prototype across different robot platforms, enabling efficient skill transfer.
		
		A major contribution of this work is the incorporation of decomposed adversarial imitation learning (DAIL), which plays a central role in overcoming the data bottleneck often encountered in humanoid robot. This decomposition not only simplifies the coordination of multi-joint systems but also significantly reduces the amount of data required to learn complex loco-manipulation tasks.
		
		Through the combination of kinematic retargeting and decomposed imitation learning, the framework allows generalized skills to be transferred across multiple humanoid platforms, while accounting for their unique configurations and dynamic properties. As a result, the same loco-manipulation tasks can be performed on diverse robots with minimal data requirements, making this approach highly effective in low-data regimes and a valuable contribution to the development of humanoid robots.
		
		\section{Discussion}\label{VI}
		
		While this work demonstrates a promising approach to cross-embodiment skill transfer, there are several areas for future improvement and exploration. First, this article has primarily focused on a common loco-manipulation task: walking to pick up and place an object. However, the framework could be extended to incorporate more advanced skills, such as force closure, energy balance, and the use of additional contact points to enhance manipulation capabilities. These advanced skills are essential for further improving the versatility and effectiveness of humanoid robots in complex loco-manipulation tasks.
		
		Another key area for future research is the reliance on full world state observation, particularly relative distances in an interaction graph. Future work could use object pose estimation to construct relative positions, reducing this dependency and improving real-world practicality.
		
		Finally, the disparity between the dynamics and motor performance of real robots compared to their simulated counterparts remains a significant challenge. While our current framework provides a solid foundation, bridging this gap is critical for the practical deployment of these skills in real-world scenarios. Future efforts could explore how human embodiment demonstrations can further guide robot learning, ensuring that the skills learned in simulation transfer effectively to physical robots.
		
		
		\bibliographystyle{ieeetr}
		\bibliography{ref_hotu}
	\end{document}